\documentclass{article} 
\PassOptionsToPackage{dvipsnames}{xcolor}
\usepackage{iclr2027_conference,times}
\usepackage[T1]{fontenc}

\usepackage{amsmath,amsfonts,bm}

\def\eqref#1{equation~\ref{#1}}

\def\1{\bm{1}}

\DeclareMathAlphabet{\mathsfit}{\encodingdefault}{\sfdefault}{m}{sl}
\SetMathAlphabet{\mathsfit}{bold}{\encodingdefault}{\sfdefault}{bx}{n}

\usepackage{amsmath,amssymb,mathtools}
\usepackage{booktabs}
\usepackage{graphicx}
\usepackage{microtype}
\usepackage{multirow}
\usepackage{xcolor}
\usepackage{hyperref}
\hypersetup{hidelinks}
\usepackage{url}
\usepackage{marvosym}

\usepackage{wrapfig}
\usepackage{pifont}
\usepackage{adjustbox}
\usepackage{subcaption}
\usepackage{array}
\usepackage{xcolor}
\usepackage{colortbl}
\usepackage{arydshln}

\newsavebox{\arcticLeftBox}
\newlength{\arcticChartGap}
\newlength{\arcticChartH}
\newlength{\arcticImageH}

\DeclareCaptionLabelFormat{figureletter}{Fig.~#2}

\definecolor{rankone}{RGB}{220,0,0}
\definecolor{ranktwo}{RGB}{0,160,0}
\definecolor{rankthree}{RGB}{0,70,210}

\newcommand{\bestresult}[1]{\textbf{#1}}
\newcommand{\secondresult}[1]{\underline{#1}}
\newcommand{\thirdresult}[1]{#1}

\usepackage{arydshln}

\newcommand{\yesmark}{\textcolor{ForestGreen}{\ding{51}}}
\newcommand{\nomark}{\textcolor{red}{\ding{55}}}

\definecolor{todored}{RGB}{180,25,25}

\newcommand{\method}{Exo2EgoHOI}

\newcommand{\patch}{\mathcal{P}}

\title{\raggedright\method: Hand-Object-Interaction Aware Exocentric-to-Egocentric Video Generation}

\author{%
\textbf{Hongjia Zhai}\textsuperscript{1},
\textbf{Xiyu Zhang}\textsuperscript{2},
\textbf{Haoran Zhang}\textsuperscript{1},
\textbf{Zhichao Ye}\textsuperscript{3},
\textbf{Haomin Liu}\textsuperscript{3},\\%
\textbf{Guofeng Zhang}\textsuperscript{2,3},
\textbf{Ian Reid}\textsuperscript{1},
\textbf{Xingxing Zuo}\textsuperscript{1,\Letter}
\\[2pt]
\textsuperscript{1}MBZUAI
\qquad
\textsuperscript{2}Zhejiang University
\qquad
\textsuperscript{3}InSpatio
}

 \iclrfinalcopy 
\begin{document}
\maketitle

\begingroup
  \renewcommand{\thefootnote}{\Letter}
  \footnotetext[0]{Corresponding author.
    Email: \texttt{xingxing.zuo@mbzuai.ac.ae}}
\endgroup

\begin{figure}[h!]
  \centering
  \includegraphics[width=0.98\linewidth]{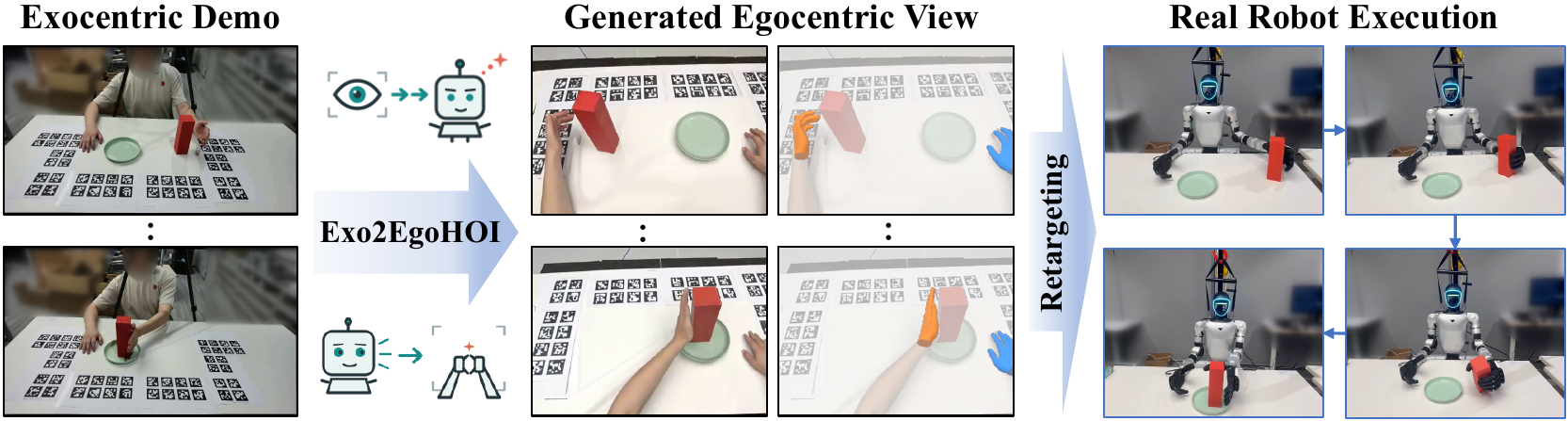}
  \vspace{-2mm}
\caption{
\textbf{From exocentric human demonstrations to egocentric observations and robot execution.}
\method{} preserves hand-object interactions during exocentric-to-egocentric video generation, enabling downstream robot execution via retargeting.}
\label{fig:teaser}
\end{figure}

\begin{abstract}
Egocentric videos of human manipulation provide valuable visual experience for embodied intelligence, yet collecting such data at scale is costly. Exocentric-to-egocentric video generation offers a scalable alternative by transforming abundant third-person manipulation videos into first-person observations. However, existing methods often struggle to faithfully preserve demonstrated hand-object interactions (HOI) across large viewpoint changes due to insufficient fine-grained interaction guidance and weak object-centric anchoring. We present \method{}, an HOI-aware video generative framework for interaction-preserving exocentric-to-egocentric translation. To preserve fine-grained HOI, we introduce a unified 4D HOI prior that combines scene geometry, articulated hand renderings, and dense hand-object relation fields, together with a dual-branch residual adapter for injecting structural and relational cues into the video generation backbone. To preserve object consistency, we introduce Decomposed Gated Cross-Attention, which separately encodes object and background references and adaptively integrates global semantic and local appearance features as object-centric anchors. Experiments on ARCTIC-HOI and Ego-Exo4D demonstrate substantial improvements in object consistency and HOI preservation while maintaining competitive visual fidelity. In particular, on ARCTIC-HOI, \method{} improves object mIoU by 32.3\% and reduces MPJPE and PA-MPJPE by 34.7\% and 50.0\%, respectively, relative to the respective best baseline results. Project page: \url{https://rcl-robotics.github.io/Exo2EgoHOI/}.
\end{abstract}

\section{Introduction}
\label{sec:introduction}

Egocentric videos of human manipulation provide valuable visual experience for embodied intelligence~\citep{humanego2026,zerowam,egoscale,egovla,lu2026world}, capturing hand movements, object changes, and actions from the actor's perspective. Such data are commonly collected using wearable cameras~\citep{ego4d,egomimic}, but scaling collection across diverse people, environments, and tasks requires substantial effort. In contrast, the Internet contains abundant third-person footage of everyday human manipulation. This motivates \emph{exocentric (source)-to-egocentric (target) video translation}, which converts readily available third-person videos into first-person visual experiences. As illustrated in Fig.~\ref{fig:teaser}, the generated egocentric observations can further provide an interface for transferring human manipulation to robotic systems. Building on recent advances in video generation and view-controllable synthesis~\citep{vista4d,ViewCrafter,wan}, recent methods have begun exploring such viewpoint translation~\citep{exo2egov,exo2egosyn,egox,park2026egoworld,groundedexo2ego}.

Despite recent progress, faithfully translating manipulation videos remains challenging: existing methods may produce visually plausible target-view videos while altering the demonstrated \emph{hand-object interactions (HOI)}, such as finger configurations or hand-object contacts. We identify two key challenges. First, \emph{scene-level geometric conditioning is insufficient to preserve fine-grained interaction geometry}. Hands and manipulated objects occupy only a small, often occluded portion of the scene, leaving fine-grained finger configurations and local hand-object relationships poorly resolved despite scene-level geometric constraints. Second, \emph{maintaining manipulated-object consistency is challenging under large exocentric-to-egocentric viewpoint changes}. Such viewpoint changes expose object regions unseen in the source view, making target-view appearance only partially constrained by geometry-based warping and rendering~\citep{ViewCrafter}. Meanwhile, global image conditioning lacks explicit object-centric correspondence, diluting object-specific cues with dominant background information. As a result, the manipulated object may drift in appearance, shape, or state throughout generation. Faithful HOI-preserving translation therefore requires both fine-grained interaction and persistent object-centric anchoring throughout cross-view video generation.

\begin{figure*}[t]
  \centering
  \includegraphics[width=\textwidth]{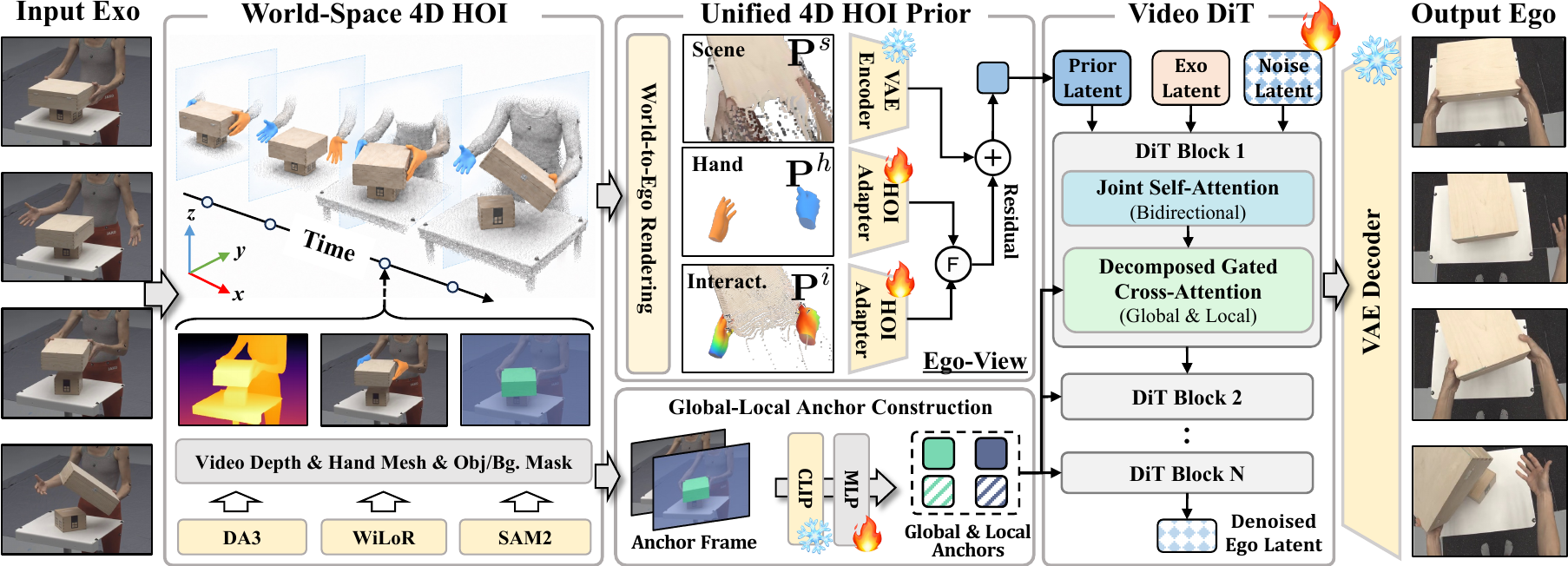}
  \vspace{-5mm}
  \caption{\textbf{Overview of \method.} Given an exocentric video, we reconstruct world-space 4D HOI and transform it into a unified egocentric prior encoding scene geometry, hand motion, and interaction cues.
HOI adapters inject this prior into a video Diffusion Transformer (DiT), while Decomposed Gated Cross-Attention anchors object-specific semantic and appearance cues from the exocentric reference to preserve object consistency throughout egocentric video generation.
}
\vspace{-3mm}
  \label{fig:overview}
\end{figure*}

To address these challenges, we introduce \method, an HOI-aware video generative framework for exocentric-to-egocentric viewpoint translation (Fig.~\ref{fig:overview}). First, we introduce a \emph{unified 4D HOI prior} that augments scene-level geometry with temporally aligned articulated hand renderings and dense hand-object relation fields in the target view. The hand renderings specify target-view hand poses and finger configurations, while the relation fields encode local proximity and relative direction between the hands and the manipulated object. A dual-branch residual HOI adapter separately encodes these structural and relational cues and fuses them into residual features, providing explicit fine-grained interaction guidance to the video generative model.
Second, we introduce \emph{Decomposed Gated Cross-Attention} to preserve object consistency across viewpoints and throughout video generation. Instead of conditioning on the source image as a single global reference, we separately encode the manipulated object and background from the first exocentric frame, providing the object with a dedicated source-view anchor. Their global semantic and local appearance features are adaptively integrated through learned per-block fusion, preserving object-specific characteristics while retaining complementary scene context.
To systematically evaluate HOI-aware viewpoint translation, we further construct an \emph{interaction-oriented evaluation benchmark} based on ARCTIC~\citep{arctic} and a collection of in-the-wild HOI videos. The benchmark jointly evaluates visual fidelity, object consistency, and hand-pose consistency, assessing both overall generation quality and preservation of the demonstrated hand-object interactions.

In summary, our main contributions are:
(i) We propose \method, an HOI-aware exocentric-to-egocentric video generation framework, featuring a \emph{unified 4D HOI prior} and dual-branch residual adapter for preserving fine-grained hand-object interactions under large viewpoint changes.
(ii) We introduce \emph{Decomposed Gated Cross-Attention}, which separately models object and background references and adaptively integrates global semantic and local appearance cues to preserve manipulated-object consistency across viewpoints and throughout video generation.
(iii) Extensive experiments on ARCTIC-HOI and Ego-Exo4D demonstrate substantial improvements over state-of-the-art methods, particularly in object consistency and interaction preservation, while maintaining strong visual fidelity. Comprehensive ablations further validate the proposed components.

\section{Related Work}
\label{sec:related}

\textbf{Exocentric-to-Egocentric Video Generation.}
Recent controllable video generation methods incorporate camera trajectories and rendered scene geometry into pretrained diffusion models to enable viewpoint and camera-motion control~\citep{cameractrl,vd3d,vista4d,ViewCrafter,inspatio_world}. Together with large-scale egocentric and synchronized ego-exo datasets~\citep{ego4d,egoexo4d}, these advances have facilitated exocentric-to-egocentric generation. Exo2Ego-V~\citep{exo2egov} and Exo2EgoSyn~\citep{exo2egosyn} exploit synchronized external views, while recent methods increasingly address single-view inputs. EgoX~\citep{egox} uses rendered scene priors and geometry-guided attention, EgoWorld~\citep{park2026egoworld} incorporates geometry and hand-pose conditioning, and Grounded-Exo2Ego~\citep{groundedexo2ego} introduces object-level semantic grounding to complement incomplete geometric evidence. Despite these advances, existing conditioning primarily focuses on scene-level alignment or coarse object grounding, while fine-grained interaction preservation and persistent object-centric anchoring remain insufficiently addressed. Our work instead targets \emph{interaction-preserving} viewpoint translation by explicitly modeling HOI structure throughout the video.

\textbf{Hand-Object Interaction Modeling and Generation.}
Accurate HOI modeling requires not only articulated hand geometry but also the spatial and contact relationships between hands and manipulated objects. Parametric hand models and recent reconstruction methods~\citep{mano,HaMeR,zeng2026stablehand,wilor} provide strong priors for recovering hand articulation, while datasets such as ARCTIC~\citep{arctic} support detailed 3D modeling of interacting hands and objects. Beyond hand pose alone, recent approaches explicitly model hand-object relations for interaction reconstruction and synthesis. GRIP~\citep{grip} exploits spatiotemporal body-object relations, GeneOH Diffusion~\citep{geneoh} introduces contact-centric representations of hand-object geometry, and DiffH2O~\citep{diffh2o} jointly models hand and object motion for interaction synthesis. More recent generative approaches~\citep{zeng2026flowhoi,egohoi,wang2026hand2world} further incorporate hand kinematics or contact structure into dynamic HOI generation. These works demonstrate the importance of explicit interaction cues beyond visual appearance or hand pose alone. In contrast, our goal is not to synthesize a new plausible interaction, but to \emph{preserve an observed interaction across viewpoints}. 

\section{Method}
\label{sec:method}
\subsection{Problem Formulation}
As shown in Fig.~\ref{fig:overview}, given an exocentric hand-object interaction video $\mathbf{X}^x=\{\mathbf{I}_t^x\}_{t=1}^{T}$, our goal is to generate its corresponding egocentric sequence $\mathbf{X}^e=\{\mathbf{I}_t^e\}_{t=1}^{T}$ while faithfully preserving the observed HOI. We condition the generation on a text description $l$, camera calibration $\mathcal{C}$, a unified ego-view \emph{4D HOI prior} $\mathbf{P}^{4D}$, and the object and background masks $(\mathbf{M}_1^{o},\mathbf{M}_1^{b})$ from the first exocentric frame. The conditional generation problem is formulated as:
\begin{equation}
  p_\theta\!\left(\mathbf{X}^{e}\mid
  \mathbf{X}^{x},l,\mathcal{C},\mathbf{P}^{4D},
  \mathbf{M}_1^{o},\mathbf{M}_1^{b}\right),
\end{equation}
where $\mathbf{P}^{4D}=\{\mathbf{P}^{s},\mathbf{P}^{h},\mathbf{P}^{i}\}$ consists of the scene, hand, and interaction priors, respectively. $\mathcal{C}$ contains the camera intrinsics and extrinsics for the exocentric view, $\mathbf{K}^{x}$ and $\{\mathbf{T}_{wc,t}^{x}\}$, and those for the target egocentric view, $\mathbf{K}^{e}$ and $\{\mathbf{T}_{wc,t}^{e}\}$.

\subsection{Unified 4D HOI Prior}
\label{sec:adapter}

Scene-level geometry provides useful spatial constraints for viewpoint translation, but does not explicitly characterize fine-grained hand-object interactions. We therefore construct a unified 4D HOI prior with three complementary components: a scene prior $\mathbf{P}^{s}$ describing the global scene layout, a hand prior $\mathbf{P}^{h}$ specifying articulated hand geometry, and an interaction prior $\mathbf{P}^{i}$ encoding local hand-object spatial relations. Together, they provide progressively finer constraints on scene structure, hand articulation, and hand-object interaction throughout the target egocentric sequence.

\textbf{World-Space 4D HOI Reconstruction.}
Given an exocentric HOI video $\mathbf{X}^{x}$, we first reconstruct the observed scene, hands, and manipulated object in a shared world coordinate system using off-the-shelf models. Specifically, DA3~\citep{depthanything3} estimates temporally consistent depth maps, WiLoR~\citep{wilor} estimates MANO~\citep{mano} hand parameters, and SAM2~\citep{sam2} provides object and background masks:
\begin{equation}
D_t^{x}=\operatorname{DA3}(\mathbf{X}^{x})_t,\quad
(\boldsymbol{\beta}_t,\boldsymbol{\theta}_t,\boldsymbol{\tau}_t)
=\operatorname{WiLoR}(\mathbf{X}^{x})_t,\quad
(\mathbf{M}_t^{o},\mathbf{M}_t^{b})=\operatorname{SAM2}(\mathbf{X}^{x})_t.
\label{eq:hoi_preprocessing}
\end{equation}
Here, $t$ indexes frames, and $\boldsymbol{\beta}_t$, $\boldsymbol{\theta}_t$, and $\boldsymbol{\tau}_t$ denote hand shape, pose, and translation, respectively. We back-project the depth maps into scene point clouds $\mathcal{P}_t^{s}$ and extract object point clouds $\mathcal{P}_t^{o}$ using the object masks $\mathbf{M}_t^{o}$. Given a pixel $\mathbf{u}=[u,v]$ in the exocentric view and its depth $D_t^{x}(u,v)$, the corresponding scene point is transformed into the shared world frame using the exocentric camera intrinsics and extrinsics:
\begin{equation}
 \mathcal{P}_t^{s}(u,v)=
 \mathbf{T}^{x}_{wc,t}
 [D_t^{x}(u,v)(\mathbf{K}^{x})^{-1}\mathbf{u};1].
 \label{eq:world_hoi_reconstruction}
\end{equation}
The resulting temporal sequence forms a \emph{world-space 4D HOI reconstruction}, from which scene, hand, and interaction cues can be consistently projected into the  egocentric view using $\mathbf{K}^{e}$, $\mathbf{T}^{e}_{wc,t}$.

\textbf{Ego-View Scene Prior.}
Similar to~\citep{ViewCrafter}, we render the scene point cloud $\mathcal{P}^{s}_t$ from the target egocentric camera at each time step and stack the rendered frames to construct the scene priors:
\begin{equation}
\mathbf{P}^{s}
=\left\{\mathcal{R}_{s}(\mathcal{P}_t^{s};\mathbf{K}^{e},\mathbf{T}_{wc,t}^{e})\right\}_{t=1}^{T}
\in\mathbb{R}^{3\times T\times H\times W},
\label{eq:scene_prior_rendering}
\end{equation}
where $\mathcal{R}_{s}$ denotes point-cloud rendering~\citep{ravi2020pytorch3d}. The scene prior provides the global layout and coarse visual context in the target view. However, because hands and manipulated objects are small and frequently occluded, scene geometry alone provides only incomplete constraints on fine-grained HOI. We therefore complement it with explicit hand and interaction priors.

\textbf{Ego-View Hand Prior.}
To explicitly constrain hand articulation, we construct the hand prior from the complete MANO hand mesh~\citep{mano}, including hand surfaces that may be occluded in the source view. Given the estimated hand shape $\boldsymbol{\beta}_t$, pose $\boldsymbol{\theta}_t$, and wrist translation $\boldsymbol{\tau}_t$, we obtain the world-space hand vertices $\mathcal{V}_t^{h}$ and mesh faces $\mathcal{F}_t^{h}$ as:
$\mathcal{V}_t^{h},\mathcal{F}_t^{h} =
\mathbf{T}^{x}_{wc,t}
\left[
\operatorname{MANO}(\boldsymbol{\beta}_t,\boldsymbol{\theta}_t)
+\boldsymbol{\tau}_t
\right].$
We then render the hand meshes with assigned RGB colors $\mathbf{C}_t$ into the target egocentric view:
\begin{equation}
\mathbf{P}^{h}
=\left\{\mathcal{R}_{h}\!\left(
\mathcal{V}_t^{h},\mathcal{F}_t^{h},\mathbf{C}_t;
\mathbf{K}^{e},\mathbf{T}_{wc,t}^{e}\right)\right\}_{t=1}^{T}
\in\mathbb{R}^{3\times T\times H\times W},
\label{eq:hand_prior_rendering}
\end{equation}
where the hand renderer $\mathcal{R}_{h}$ uses a hand-only $z$-buffer and barycentric interpolation to produce per-pixel colors. The resulting prior explicitly specifies target-view hand shape and articulation beyond what can be recovered from the partial scene geometry.

\textbf{Ego-View Interaction Prior.}
While $\mathbf{P}^{h}$ specifies hand articulation, it does not explicitly describe how the hand interacts with the manipulated object. We therefore characterize the local hand-object relation at each hand vertex using three complementary quantities: contact likelihood, hand-object distance, and relative direction.
For each world-space hand vertex $\mathbf{v}^{w}_{t,i}\in\mathcal{V}^{h}_t$, we identify its five nearest object points in $\mathbf{p}^{o}_{t,j} \in \mathcal{P}^{o}_t$ and denote their index set by $\mathcal{N}_{t,i}$. The distance to each neighboring object point is
$\delta_{t,i,j}=\|\mathbf{v}^{w}_{t,i}-\mathbf{p}^{o}_{t,j}\|_2$.
We assign normalized inverse-distance weights and compute the weighted mean distance, where $\epsilon$ ensures numerical stability:
\begin{equation}
a_{t,i,j}=\max(\delta_{t,i,j},\epsilon)^{-1}\Big/\sum_{j'\in\mathcal{N}_{t,i}}\max(\delta_{t,i,j'},\epsilon)^{-1},\quad d_{t,i}=\sum_{j\in\mathcal{N}_{t,i}}a_{t,i,j}\delta_{t,i,j}.
\label{eq:nearest_relation}
\end{equation}
Using the same weights, we compute the displacement from the hand vertex to the weighted center of its neighboring object points:
$\mathbf{q}^{w}_{t,i}=\sum_{j\in\mathcal{N}_{t,i}}a_{t,i,j}\mathbf{p}^{w}_{t,j}-\mathbf{v}^{w}_{t,i}$.
We then encode the local relation using a soft contact score $s_{t,i}$, normalized distance $\bar d_{t,i}$, and interaction direction $\mathbf{r}^{w}_{t,i}$:
\begin{equation}
s_{t,i}
=
\exp\!\left(-d_{t,i}^{2}/(2\sigma^{2})\right),\quad
\bar d_{t,i}
=
\operatorname{clip}\!\left(d_{t,i}/d_{\max},0,1\right),\quad
\mathbf{r}^{w}_{t,i}
=
\frac{\mathbf{q}^{w}_{t,i}}
{\max(\|\mathbf{q}^{w}_{t,i}\|_2,\epsilon)}.
\label{eq:contact_distance}
\end{equation}
Here, $\sigma$ controls the contact-score bandwidth, $d_{\max}$ sets the distance normalization range. 

Using the rotation component $\mathbf{R}^{e}_{cw,t}$ of $\mathbf{T}^{e}_{cw,t}$, we transform the interaction direction into the target egocentric camera frame:
$\mathbf{r}^{e}_{t,i}=\mathbf{R}^{e}_{cw,t}\mathbf{r}^{w}_{t,i}$.
Each hand vertex is therefore associated with a five-dimensional descriptor
$[s_{t,i},\bar d_{t,i},\mathbf{r}^{e}_{t,i}]\in\mathbb{R}^{5}$.
Let $\mathcal{A}_t$ collect these descriptors over all hand vertices. We rasterize them into target egocentric view using the same hand topology as Eq.~\ref{eq:hand_prior_rendering}:
\begin{equation}
  \mathbf{P}^{i}
  =\left\{\mathcal{R}_h\!\left(
  \mathcal{V}^{h}_{t},\mathcal{F}_t^{h},\mathcal{A}_t;
  \mathbf{K}^{e},\mathbf{T}^{e}_{wc,t}\right)\right\}_{t=1}^{T}
  \in\mathbb{R}^{5\times T\times H\times W}.
  \label{eq:relation_rasterization}
\end{equation}
The resulting $\mathbf{P}^{i}$ is pixel-aligned with $\mathbf{P}^{s}$ and $\mathbf{P}^{h}$, providing explicit target-view hand-object relational cues throughout the sequence.

\textbf{Dual-Branch HOI Residual Adapter.}
We next integrate the complementary priors into the video generation backbone. The scene prior $\mathbf{P}^{s}$ is encoded by the frozen VAE of Wan~\citep{wan}, followed by patch embedding to obtain scene-conditioned egocentric tokens $\mathbf{Z}_{\mathrm{in}}^{e}$. Since the RGB-like hand rendering $\mathbf{P}^{h}$ and the non-photometric relation field $\mathbf{P}^{i}$ have different signal characteristics, we encode them using two independent spatiotemporal branches, $\mathcal{F}_h$ and $\mathcal{F}_i$. Each branch consists of six 3D convolutional layers with SiLU activations, progressively downsampling the spatiotemporal resolution to match the video latent space. The two branches share the same architecture but use independent parameters, with input channels adapted to the 3-channel hand prior and 5-channel interaction prior, respectively. Detailed architecture is provided in the supplementary material.

The resulting feature grids are concatenated along the channel dimension, fused using a $1{\times}1{\times}1$ convolution $\mathcal{F}_{\mathrm{fuse}}$, and projected by $\patch$ to match the spatiotemporal resolution and channel dimension of $\mathbf{Z}_{\mathrm{in}}^{e}$:
\begin{equation}
\Delta\mathbf{Z}^{\mathrm{hoi}}
=
\patch\!\left(
\mathcal{F}_{\mathrm{fuse}}\!\left(
[\mathcal{F}_h(\mathbf{P}^{h});
 \mathcal{F}_i(\mathbf{P}^{i})]
\right)
\right).
\label{eq:prior_tokens}
\end{equation}
The projection $\patch$ is zero-initialized such that the adapter initially produces a zero residual. We inject the fused HOI features into the scene-conditioned tokens through residual addition:
\begin{equation}
\mathbf{Z}^{4D}
=
\mathbf{Z}_{\mathrm{in}}^{e}
+
\lambda_{\mathrm{hoi}}\Delta\mathbf{Z}^{\mathrm{hoi}},
\label{eq:prior_inject}
\end{equation}
where $\lambda_{\mathrm{hoi}}$ is a learned residual scale initialized to one. The resulting HOI-conditioned tokens $\mathbf{Z}^{4D}$ integrate scene context, articulated hand structure, and explicit hand-object relations in a unified target-view representation.

\subsection{Decomposed Gated Cross-Attention}
\label{sec:crossattention}
\begin{wrapfigure}{r}{0.5\textwidth} 
  \centering
  \vspace{-1em}             
    \includegraphics[width=\linewidth]{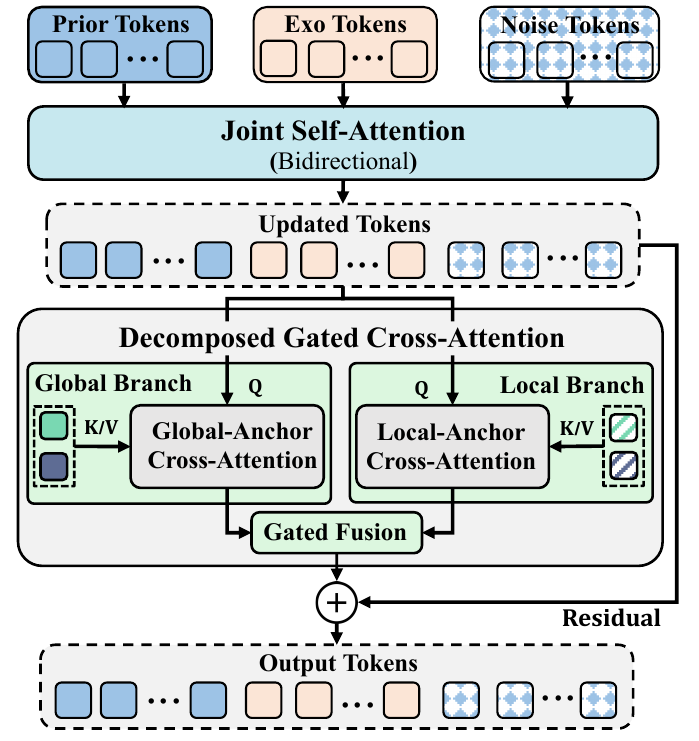}
    \vspace{-6mm}
    \caption{\footnotesize \textbf{Architecture of DiT block.} DGCA adaptively fuses global and local contexts through learnable block-specific gates.}%
    \label{fig:dit_block}
  \vspace{-4mm}             
\end{wrapfigure}
Although the unified 4D HOI prior $\mathbf{P}^{4D}$ explicitly constrains interaction geometry, it does not guarantee that the manipulated object remains consistent across large viewpoint changes and throughout the generated sequence. Geometry-based rendering provides only partial target-view evidence, as previously unseen object surfaces cannot be recovered from the source observation. Moreover, conventional full-image conditioning can underrepresent the manipulated object, which typically occupies only a small region against a dominant background. We therefore introduce \emph{Decomposed Gated Cross-Attention (DGCA)} to provide explicit object-centric anchoring during generation. DGCA separately encodes object and background references and adaptively integrates their global semantic and local appearance features through learned block-specific gates.

\textbf{Decomposed Global-Local Anchors.}
As illustrated in Fig.~\ref{fig:overview}, given the first exocentric frame $\mathbf{I}_1^x$ and its object and background masks $(\mathbf{M}_1^o,\mathbf{M}_1^b)$ from Eq.~\ref{eq:hoi_preprocessing}, we extract an object crop $\mathbf{I}_1^{x,o}$ and construct a background reference $\mathbf{I}_1^{x,b}$ with the manipulated object masked out. This regional decomposition provides the manipulated object with a dedicated representation while retaining complementary scene context.

We encode both references, $\mathbf{I}_1^{x,b}$ and $\mathbf{I}_1^{x,o}$, using a shared frozen CLIP encoder. For each region $r\in\{o,b\}$, the encoder produces a global \texttt{[CLS]} token $\mathbf{f}_{\mathrm{cls}}^r$ and spatially resolved patch tokens $\mathbf{F}_{\mathrm{patch}}^r$ via
$[\mathbf{f}_{\mathrm{cls}}^r;
 \mathbf{F}_{\mathrm{patch}}^r]=\mathcal{E}_{\mathrm{CLIP}}(\mathbf{I}_1^{x,r}), r \in \{o,b\}$. We then construct global and local reference anchors by concatenating the corresponding object and background features:
\begin{equation}
\mathbf{C}^{g}
=
[\mathbf{f}_{\mathrm{cls}}^{o};
 \mathbf{f}_{\mathrm{cls}}^{b}],
\qquad
\mathbf{C}^{\ell}
=
[\mathbf{F}_{\mathrm{patch}}^{o};
 \mathbf{F}_{\mathrm{patch}}^{b}],
\label{eq:global_reference}
\end{equation}
The resulting representation is therefore decomposed along two complementary axes: \emph{region}, separating the manipulated object from the background, and \emph{feature granularity}, separating global semantics from local appearance. The global anchor $\mathbf{C}^{g}$ captures object and scene semantics, whereas the local anchor $\mathbf{C}^{\ell}$ retains spatially resolved appearance details. Together, they provide complementary cues for anchoring object identity and appearance throughout the video generation.

\textbf{Gated Cross-Attention Fusion.}
As illustrated in Fig.~\ref{fig:dit_block}, within each DiT block, bidirectional joint self-attention first processes exocentric video tokens, noisy egocentric tokens, and HOI-conditioned tokens. The resulting video features provide queries $\mathbf{Q}_i$ for reference cross-attention in the $i$-th block. The global and local anchors are passed through the shared image projector and the block-specific image key/value projections to obtain $(\mathbf{K}_i^r,\mathbf{V}_i^r)$ for $r\in\{g,\ell\}$. The corresponding cross-attention with $ d$-dimensional attention heads is:
\begin{equation}
\mathbf{A}_i^{r}
=
\operatorname{Attn}(\mathbf{Q}_i,\mathbf{K}_i^{r},\mathbf{V}_i^{r})
=
\operatorname{softmax}\!\left(
\mathbf{Q}_i(\mathbf{K}_i^{r})^\top/{\sqrt{d}}
\right)\mathbf{V}_i^{r},
\quad r\in\{g,\ell\},
\label{eq:image_cross_attention}
\end{equation}
Rather than combining global and local appearance cues with fixed weights, we allow each DiT block to learn its own balance between the two feature granularities. Specifically, the $i$-th block learns a two-dimensional logit vector $\mathbf{a}_i$, which is normalized into gating coefficients:
\begin{equation}
\boldsymbol{\pi}_i
=
\operatorname{softmax}(\mathbf{a}_i),
\quad
\mathbf{A}_i^{\mathrm{img}}
=
\pi_i^{g}\mathbf{A}_i^{g}
+
\pi_i^{\ell}\mathbf{A}_i^{\ell}.
\label{eq:gating}
\end{equation}
Here, $\pi_i^{g}$ and $\pi_i^{\ell}$ control the contributions of global semantic and local appearance features, respectively. The gates are block-specific but shared across inputs and denoising steps, allowing different stages of the backbone to learn different global-local balances while the attention responses remain conditioned on the current video queries. The fused reference feature $\mathbf{A}_i^{\mathrm{img}}$ is incorporated into the block's cross-attention update before the shared output projection and residual addition. In this way, DGCA complements the geometric 4D HOI prior with persistent object-centric anchoring, helping preserve object identity and appearance across viewpoints and throughout the generated sequence.

\textbf{Training Objective.}
We train the model using conditional flow matching~\citep{lipman2023flow}. Given a paired egocentric training video $\mathbf{X}^{e}$, a frozen video VAE encodes it into a clean latent $\mathbf{z}_0^{e}=\mathcal{E}_{\mathrm{VAE}}(\mathbf{X}^{e})$. Let $\mathcal{K}$ collectively denote all conditioning inputs, including text prompt $l$, exocentric video context $\mathbf{X}^{x}$, unified 4D HOI prior $\mathbf{P}^{4D}$, and decomposed image reference anchors $\mathbf{C}^{\ell}$, $\mathbf{C}^{g}$.
We sample a flow-matching timestep $\tau$ from the pretrained scheduler, obtain its corresponding noise level $\sigma_\tau\in[0,1]$, and sample Gaussian noise $\boldsymbol{\epsilon}\sim\mathcal{N}(\mathbf{0},\mathbf{I})$. The noisy latent is interpolated between the clean video latent and noise, and the model is trained to predict the corresponding latent velocity:
\begin{equation}
\mathbf{z}_\tau^{e}
=
(1-\sigma_\tau)\mathbf{z}_0^{e}
+
\sigma_\tau\boldsymbol{\epsilon},
\qquad
\mathcal{L}_{\mathrm{FM}}
=
\mathbb{E}_{(\mathbf{X}^{e},\mathcal{K}),
             \tau,\boldsymbol{\epsilon}}
\left[
\left\|
\mathbf{v}_{\theta}
(\mathbf{z}_\tau^{e},\tau,\mathcal{K})
-
(\boldsymbol{\epsilon}-\mathbf{z}_0^{e})
\right\|_2^2
\right].
\label{eq:loss}
\end{equation}

\section{Experiments}
\label{sec:experiments}

\subsection{Experimental Settings}
\textbf{Datasets.}
We evaluate on Ego-Exo4D~\citep{egoexo4d} using a setup similar to EgoX~\citep{egox}, covering diverse scenes and manipulation activities. Since Ego-Exo4D lacks a dedicated evaluation protocol for HOI-preserving video generation, we additionally construct ARCTIC-HOI, an interaction-oriented benchmark comprising 7K training clips and 180 non-overlapping test clips from ARCTIC~\citep{arctic}, together with 60 in-the-wild HOI test clips.

\textbf{Evaluation Metrics.}
We evaluate three aspects of the generated egocentric videos.
For \textit{visual fidelity}, we report PSNR, SSIM, LPIPS, and CLIP
image similarity (CLIP-I).
For \textit{object consistency}, we extract object masks using SAM~2~\citep{sam2}, and report mask IoU (mIoU), normalized object center error (Center-Err.), and object-crop CLIP similarity (CLIP-O).
For \textit{hand consistency}, we extract hand using~\citep{wilor} and report MPJPE, per-frame
Procrustes-aligned MPJPE (PA-MPJPE), and sequence-level world-aligned
MPJPE (WA-MPJPE) in millimeters.
Detailed definitions of the metrics are provided in the supplementary materials.

\textbf{Implementation Details.}
We initialize the video DiT from Wan2.1-I2V-14B~\citep{wan}, which consists of 40 blocks of width 5120.
We fine-tuned the model using LoRA~\cite{lora}, rank=256, on 8 H200 GPUs, with the AdamW optimizer, learning rate $2\times10^{-5}$, betas $(0.9,0.95)$, and weight decay $10^{-4}$. We jointly optimize the LoRA parameters, the dual-branch HOI residual adapter, and the cross-attention gates, while keeping the remaining pretrained parameters frozen.
Full architecture and configuration details are given in the supplementary materials.

\textbf{Baselines.}
We compare our approach with recent exocentric-to-egocentric video generation methods, including EgoWorld~\citep{park2026egoworld} and EgoX~\citep{egox}, as well as general view-controllable video generation baselines, WAN VACE~\citep{vace} and Vista4D~\citep{vista4d}.
Since Vista4D provides inference-only code and EgoWorld does not include the complete training pipeline required for fine-tuning, we evaluate both methods using their publicly available inference implementations. For WAN VACE and EgoX, we fine-tune the released models using the same training configuration as ours for a fair comparison.

\begin{table*}[t]
\caption{
\textbf{Quantitative comparison on ARCTIC-HOI and Ego-Exo4D datasets.}
We evaluate visual fidelity, object consistency, and hand-pose
consistency. Best results are shown in \textbf{bold} and second-best results are \underline{underlined}.
}
\vspace{-2mm}
\centering
\small
\setlength{\tabcolsep}{2.8pt}
\renewcommand{\arraystretch}{1.08}

\resizebox{0.999\textwidth}{!}{%
\begin{tabular}{lcccccccccc}
\toprule
& \multicolumn{4}{c}{\textbf{Visual Fidelity}}
& \multicolumn{3}{c}{\textbf{Object Consistency}}
& \multicolumn{3}{c}{\textbf{Hand Consistency}} \\
\cmidrule(lr){2-5}
\cmidrule(lr){6-8}
\cmidrule(lr){9-11}

\textbf{Method}
& PSNR$\uparrow$
& SSIM$\uparrow$
& LPIPS$\downarrow$
& CLIP-I$\uparrow$
& mIoU$\uparrow$
& Center-Err.$\downarrow$
& CLIP-O$\uparrow$
& MPJPE$\downarrow$
& WA-MPJPE$\downarrow$
& PA-MPJPE$\downarrow$ \\
\midrule

\multicolumn{11}{c}{%
    \cellcolor{gray!15}\textbf{ARCTIC-HOI}%
} \\

WAN VACE~\citep{vace}
& \thirdresult{12.78}
& \thirdresult{0.6163}
& 0.5115
& 0.8485
& 0.2336
& \thirdresult{0.0893}
& 0.8080
& 2870.22
& \thirdresult{371.18}
& \secondresult{20.36} \\

EgoWorld~\citep{park2026egoworld}
& 11.90
& 0.4652
& 0.6203
& 0.7995
& 0.2241
& 0.1131
& 0.7869
& 2282.03
& 440.60
& 57.92 \\

Vista4D~\citep{vista4d}
& \bestresult{13.58}
& \secondresult{0.6614}
& \secondresult{0.4324}
& \thirdresult{0.8737}
& \thirdresult{0.2768}
& 0.0929
& \thirdresult{0.8176}
& \thirdresult{2123.78}
& 387.73
& 28.91 \\

EgoX~\citep{egox}
& 11.83
& 0.5941
& \thirdresult{0.4743}
& \secondresult{0.8910}
& \secondresult{0.2933}
& \secondresult{0.0804}
& \secondresult{0.8345}
& \secondresult{2019.81}
& \secondresult{325.56}
& \thirdresult{22.12} \\

Ours
& \secondresult{13.43}
& \bestresult{0.6685}
& \bestresult{0.3860}
& \bestresult{0.9175}
& \bestresult{0.3881}
& \bestresult{0.0583}
& \bestresult{0.8557}
& \bestresult{1318.74}
& \bestresult{313.75}
& \bestresult{10.18} \\

\midrule

\multicolumn{11}{c}{%
    \cellcolor{gray!15}\textbf{Ego-Exo4D}%
} \\

WAN VACE~\citep{vace}
& \thirdresult{12.67}
& \thirdresult{0.4161}
& \thirdresult{0.5981}
& \thirdresult{0.8251}
& \thirdresult{0.0362}
& \thirdresult{0.2214}
& 0.8003
& \thirdresult{6810.62}
& \thirdresult{2163.18}
& \bestresult{74.00} \\

EgoWorld~\citep{park2026egoworld}
& 11.05
& 0.3410
& 0.7186
& 0.7220
& 0.0227
& 0.2349
& 0.7911
& --
& --
& -- \\

Vista4D~\citep{vista4d}
& 10.46
& 0.2684
& 0.7348
& 0.7459
& 0.0154
& 0.2571
& \thirdresult{0.8067}
& 20074.43
& 2293.86
& 120.68 \\

EgoX~\citep{egox}
& \secondresult{13.14}
& \secondresult{0.4474}
& \secondresult{0.5258}
& \secondresult{0.8563}
& \secondresult{0.0703}
& \secondresult{0.1842}
& \secondresult{0.8298}
& \secondresult{5773.68}
& \secondresult{1928.07}
& \secondresult{75.55} \\

Ours
& \bestresult{14.11}
& \bestresult{0.4690}
& \bestresult{0.5035}
& \bestresult{0.8665}
& \bestresult{0.0723}
& \bestresult{0.1745}
& \bestresult{0.8341}
& \bestresult{5208.17}
& \bestresult{1749.36}
& \thirdresult{93.81} \\

\bottomrule
\end{tabular}%
}
\label{tab:main_results}
\vspace{-3mm}
\end{table*}
\subsection{Main Results}
We evaluate exocentric-to-egocentric video generation on ARCTIC-HOI and Ego-Exo4D, with quantitative results reported in Tab.~\ref{tab:main_results} and qualitative comparisons shown in Fig.~\ref{fig:main_results}.

\textbf{Quantitative Comparison.}
As shown in Tab.~\ref{tab:main_results}, \method{} consistently improves visual fidelity, object consistency, and hand consistency on ARCTIC-HOI. Compared with the strongest baselines, LPIPS decreases from 0.4324 to 0.3860, while object mIoU increases from 0.2933 to 0.3881. More importantly, MPJPE decreases from 2019.81 to 1318.74\,mm and PA-MPJPE from 20.36 to 10.18\,mm, corresponding to reductions of $34.7\%$ and $50.0\%$, respectively. These improvements indicate that our explicit HOI conditioning better preserves both manipulated objects and hand configurations across viewpoints. On Ego-Exo4D, \method{} similarly improves most visual and object metrics, increasing PSNR from 13.14 to 14.11\,dB and improving all three object-consistency metrics over EgoX. MPJPE and WA-MPJPE are also reduced to 5208.17 and 1749.36\,mm, respectively, although PA-MPJPE remains higher than WAN VACE and EgoX. Overall, these results demonstrate substantial gains in object consistency and HOI preservation while maintaining strong visual fidelity across both datasets. EgoWorld's hand metrics are marked as ``--'' because its hand estimator component, WiLoR~\citep{wilor}, could not reliably recover hand meshes from its generated videos.

\begin{table}[t]
\caption{
\textbf{Ablation studies on the ARCTIC-HOI dataset.}
``\#8 Vanilla CA'' uses CLIP to encode the entire anchor frame for cross-attention without decomposition.
Best results are shown in \textbf{bold} and second-best results are \underline{underlined}.
}
\vspace{-2mm}
\label{tab:prior_ablation}
\centering
\small
\setlength{\tabcolsep}{2.2pt}
\renewcommand{\arraystretch}{1.08}

\resizebox{\textwidth}{!}{%
\begin{tabular}{@{}l*{13}{c}@{}}
\toprule
\multirow{2}{*}{\textbf{Methods}}
& \multicolumn{3}{c}{\textbf{HOI Priors}}
& \multicolumn{3}{c}{\textbf{DGCA}}
& \multicolumn{7}{c}{\textbf{Evaluation Metrics}} \\
\cmidrule(lr){2-4}
\cmidrule(lr){5-7}
\cmidrule(lr){8-14}

& \textbf{Scene}
& \textbf{Hand}
& \textbf{Inter.}
& \textbf{Global}
& \textbf{Local}
& \textbf{Gate}
& LPIPS$\downarrow$
& CLIP-I$\uparrow$
& mIoU$\uparrow$
& CLIP-O$\uparrow$
& MPJPE$\downarrow$
& WA-MPJPE$\downarrow$
& PA-MPJPE$\downarrow$ \\
\midrule

\#0 Baseline
& \nomark & \nomark & \nomark
& \nomark & \nomark & \nomark
& 0.5577
& 0.8253
& 0.1148
& 0.8056
& 2631.66
& 371.25
& 41.42 \\
\hdashline

\#1 \textit{w/o} All Priors
& \nomark & \nomark & \nomark
& \yesmark & \yesmark & \yesmark
& 0.5351
& 0.8449
& 0.1378
& 0.8255
& 2254.26
& 346.98
& 21.11 \\

\#2 \textit{w/} Scene
& \yesmark & \nomark & \nomark
& \yesmark & \yesmark & \yesmark
& 0.4435
& 0.9008
& 0.3104
& 0.8391
& 1978.57
& 327.60
& 14.30 \\

\#3 \textit{w/} Scene \& Hand
& \yesmark & \yesmark & \nomark
& \yesmark & \yesmark & \yesmark
& 0.4063
& 0.9108
& 0.3323
& 0.8421
& \secondresult{1570.51}
& \secondresult{315.28}
& 13.33 \\
\hdashline

\#4 \textit{w/o} Decomp. CA
& \yesmark & \yesmark & \yesmark
& \nomark & \nomark & \nomark
& \secondresult{0.3953}
& 0.8853
& 0.3023
& 0.8198
& 2327.34
& 398.94
& 16.02 \\

\#5 \textit{w/} Global
& \yesmark & \yesmark & \yesmark
& \yesmark & \nomark & \nomark
& 0.4258
& 0.9057
& 0.3245
& 0.8405
& 1805.71
& 323.42
& 14.91 \\

\#6 \textit{w/} Local
& \yesmark & \yesmark & \yesmark
& \nomark & \yesmark & \nomark
& 0.4326
& 0.9025
& \thirdresult{0.3559}
& 0.8464
& 1845.20
& 321.22
& \thirdresult{13.30} \\

\#7 \textit{w/o} Gate
& \yesmark & \yesmark & \yesmark
& \yesmark & \yesmark & \nomark
& \thirdresult{0.4005}
& \secondresult{0.9156}
& 0.3354
& \thirdresult{0.8479}
& 1638.83
& 322.44
& 13.61 \\

\#8 Vanilla CA
& \yesmark & \yesmark & \yesmark
& -- & -- & --
& 0.4167
& \thirdresult{0.9108}
& \secondresult{0.3618}
& \secondresult{0.8488}
& \thirdresult{1598.81}
& \thirdresult{320.69}
& \secondresult{11.39} \\
\hdashline

\#9 Ours (Full)
& \yesmark & \yesmark & \yesmark
& \yesmark & \yesmark & \yesmark
& \bestresult{0.3860}
& \bestresult{0.9175}
& \bestresult{0.3881}
& \bestresult{0.8557}
& \bestresult{1318.74}
& \bestresult{313.75}
& \bestresult{10.18} \\

\bottomrule
\end{tabular}%
}
\end{table}

\begin{figure*}[h]
    \centering
    \scriptsize
    \setlength{\tabcolsep}{0pt}
    \renewcommand{\arraystretch}{1.0}

    \begin{tabular}{@{}c*{6}{c}@{}}
        \multicolumn{7}{@{}c@{}}{%
            \includegraphics[
                page=1,
                pagebox=cropbox,
                width=\textwidth
            ]{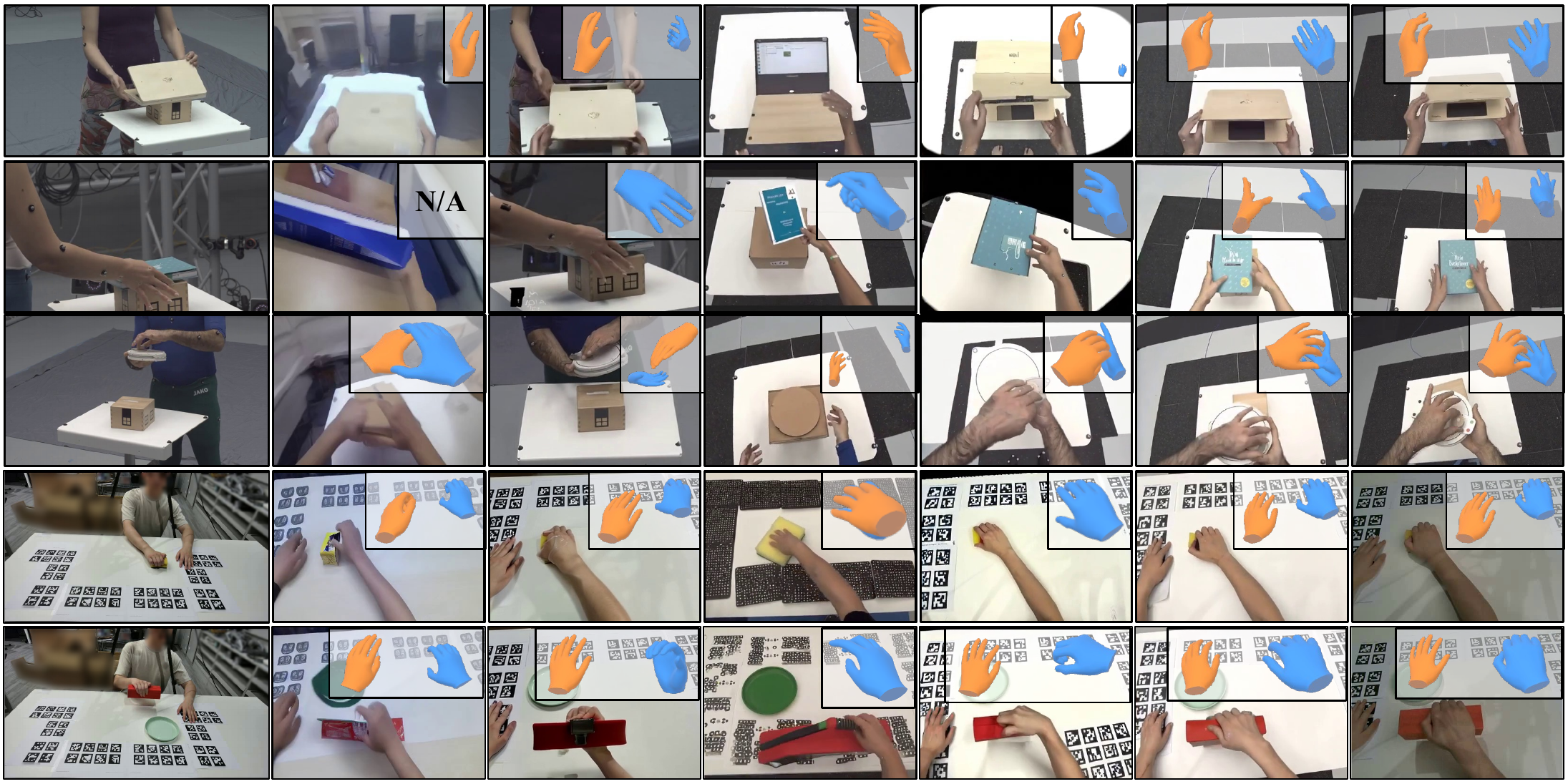}%
        } \\[0pt]

        \makebox[0.172414\textwidth][c]{\textbf{Exocentric View}} &
        \makebox[0.137931\textwidth][c]{\textbf{EgoWorld}} &
        \makebox[0.137931\textwidth][c]{\textbf{Vista4D}} &
        \makebox[0.137931\textwidth][c]{\textbf{WAN VACE}} &
        \makebox[0.137931\textwidth][c]{\textbf{EgoX}} &
        \makebox[0.137931\textwidth][c]{\textbf{Ours}} &
        \makebox[0.137931\textwidth][c]{\textbf{Ego GT}} \\
    \end{tabular}
    \vspace{-3mm}
    \caption{
        \textbf{Qualitative comparison on ARCTIC-HOI dataset.}
        Compared with existing methods, \method{} better preserves the target-view object appearance and hand-object configuration under large viewpoint changes. Insets visualize hand meshes recovered from the generated egocentric frames for comparison with the reference.
    }
    \label{fig:main_results}
    \vspace{-4mm}
\end{figure*}

\textbf{Qualitative Comparison.}
Fig.~\ref{fig:main_results} compares generated egocentric views together with their estimated hand meshes. Existing methods can produce plausible egocentric results but often alter object appearance, placement, or hand configuration under large viewpoint changes. In the hinged-object example, our generated video better preserves the raised lid and circular interior of the target object, while in the scissors example ours more closely reproduces the target-view tool placement and bimanual configuration. Similar improvements are observed in the remaining examples, where \method{} better maintains object consistency and HOI configurations during video generation.

\subsection{Ablation Studies}
\label{sec:ablation}

\begin{figure*}[t]
    \centering

    \setlength{\arcticChartGap}{3pt}

    \sbox{\arcticLeftBox}{%
        \includegraphics[
            page=1,
            pagebox=cropbox,
            width=0.70\textwidth
        ]{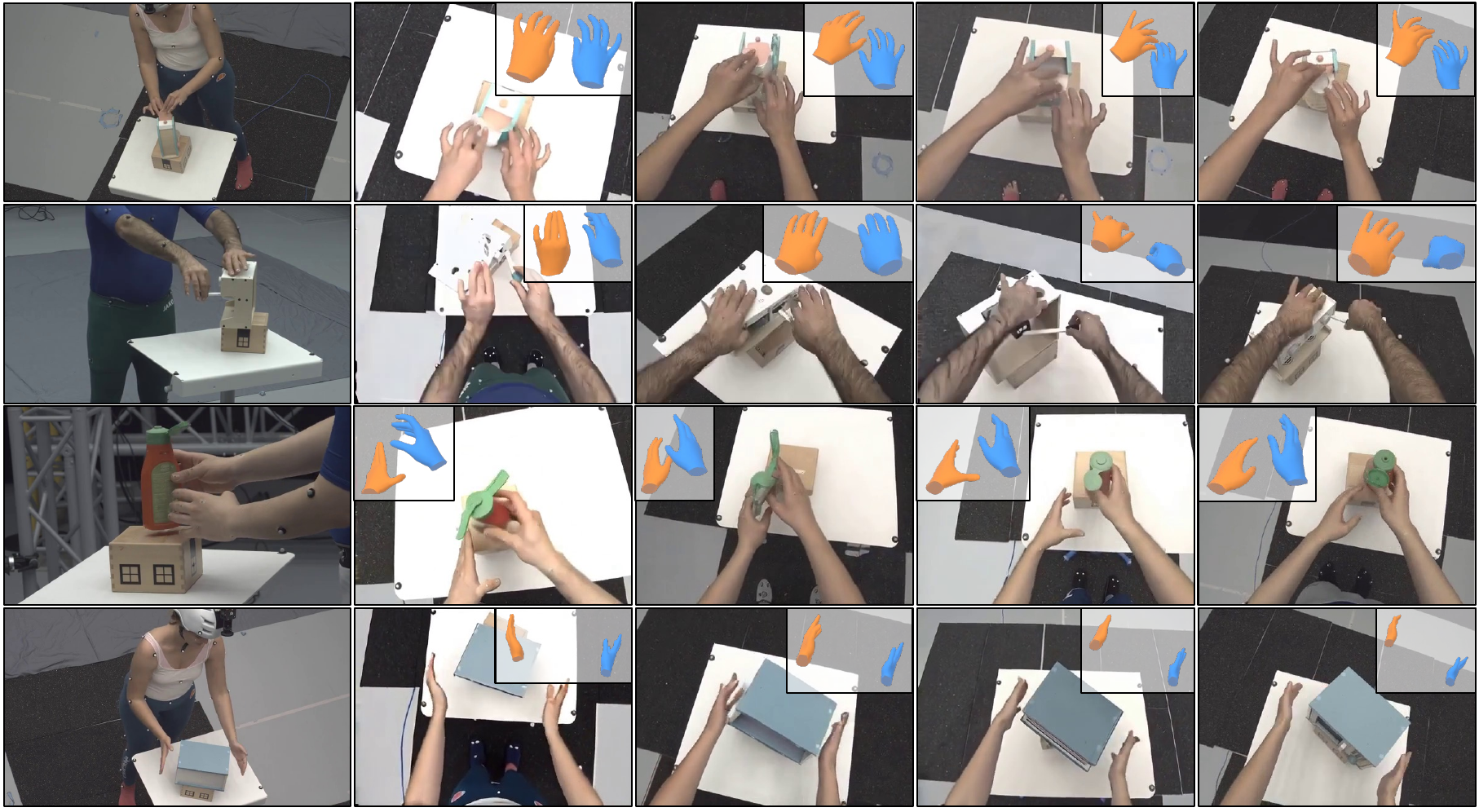}%
    }

    \setlength{\arcticImageH}{%
        \dimexpr\ht\arcticLeftBox+\dp\arcticLeftBox\relax
    }

    \setlength{\arcticChartH}{%
        \dimexpr(\arcticImageH-\arcticChartGap)/2\relax
    }

    \begin{minipage}[t]{0.70\textwidth}
        \vspace{0pt}
        \centering

        \usebox{\arcticLeftBox}

        {\scriptsize
            \noindent
            \makebox[\linewidth][c]{%
                \strut
                \makebox[0.238095\linewidth][c]{\textbf{Exocentric View}}%
                \makebox[0.190476\linewidth][c]{%
                    \textbf{\#1 \textit{w/o} Priors}}%
                \makebox[0.190476\linewidth][c]{%
                    \textbf{\#4 \textit{w/o} DGCA}}%
               \makebox[0.190476\linewidth][c]{\textbf{Ours}}%
               \makebox[0.190476\linewidth][c]{\textbf{Ego GT}}%
            }%
            \par
        }
    \end{minipage}%
    \hfill
    \begin{minipage}[t]{0.4pt}
        \vspace{0pt}
        \noindent
        \rule{0.4pt}{\arcticImageH}
    \end{minipage}%
    \hfill
    \begin{minipage}[t]{0.28\textwidth}
        \vspace{0pt}
        \centering

        \makebox[\linewidth][c]{%
            \vbox{%
                \offinterlineskip
                \hbox{%
                    \includegraphics[
                        pagebox=cropbox,
                        height=\arcticChartH
                    ]{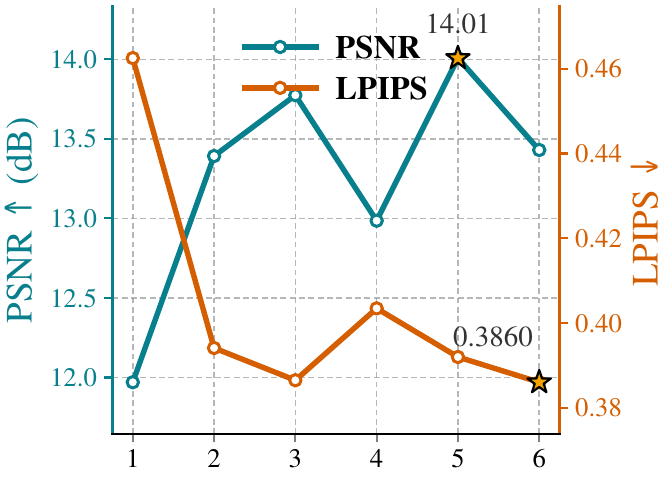}%
                }%
                \vskip\arcticChartGap
                \hbox{%
                    \includegraphics[
                        pagebox=cropbox,
                        height=\arcticChartH
                    ]{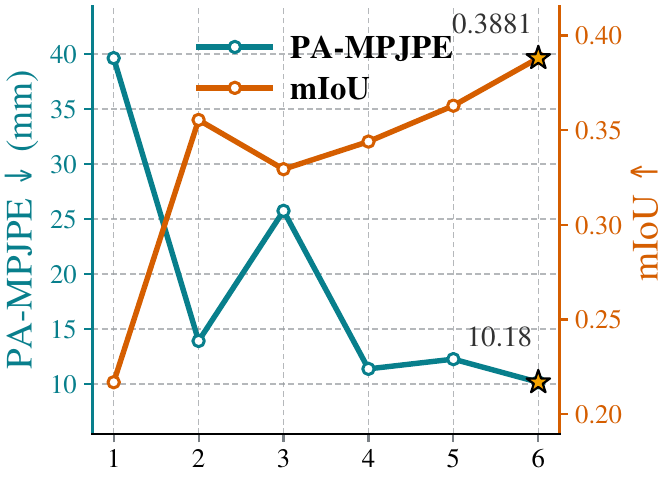}%
                }%
            }%
        }%
        \par
        {\scriptsize
            \strut \textbf{Number of Training Views}\par
        }
    \end{minipage}
    \vspace{-2mm}
    \caption{
        \textbf{Ablation studies on the ARCTIC-HOI dataset.}
        Left: qualitative comparison of different configurations.
        Right: visual fidelity (PSNR and LPIPS; top) and hand/object consistency (PA-MPJPE and mIoU; bottom) with different numbers of exocentric training views.
    }
    \label{fig:arctic_combined_ablation}
    \vspace{-4mm}
\end{figure*}

We ablate the key designs of \method{} on ARCTIC-HOI in Tab.~\ref{tab:prior_ablation}. Fig.~\ref{fig:arctic_combined_ablation} provides qualitative comparisons and analyzes the effect of using different numbers of exocentric training views.

\textbf{Unified 4D HOI Prior.}
We first study the complementary roles of scene, hand, and interaction priors. Without geometric priors (\#1), the model obtains an mIoU of 0.1378 and MPJPE of 2254.26\,mm. Introducing the scene prior (\#2) substantially increases mIoU to 0.3104 and reduces MPJPE to 1978.57\,mm, demonstrating the importance of target-view scene geometry. Adding the articulated hand prior (\#3) further reduces MPJPE to 1570.51\,mm and PA-MPJPE from 14.30 to 13.33\,mm, showing that explicit hand geometry provides finer constraints beyond scene-level conditioning. Finally, adding the interaction prior while keeping DGCA unchanged (\#3$\rightarrow$\#9) increases mIoU from 0.3323 to 0.3881 and reduces MPJPE and PA-MPJPE to 1318.74 and 10.18\,mm, respectively. These results show that scene layout, hand articulation, and hand-object relations provide complementary levels of geometric guidance, with explicit relational cues further improving both object and hand consistency.

\textbf{Decomposed Gated Cross-Attention.}
We next analyze the components of DGCA for preserving object consistency while retaining all HOI priors (\#4--\#9). Compared with no DGCA (\#4), introducing either the global (\#5) or local (\#6) branch improves object and hand consistency. The global branch achieves higher CLIP-I, while the local branch provides better mIoU and PA-MPJPE, supporting their complementary roles in providing semantic and fine-grained appearance anchors for object consistency. Combining both branches with fixed fusion (\#7) further improves CLIP-I and MPJPE, while replacing fixed weights with learned per-block gates (\#9) increases mIoU from 0.3354 to 0.3881 and reduces PA-MPJPE from 13.61 to 10.18\,mm. Moreover, using global and local features without object-background decomposition (\#8) degrades mIoU from 0.3881 to 0.3618 and PA-MPJPE from 10.18 to 11.39\,mm. Together, these results validate both regional decomposition and adaptive global-local fusion in DGCA. Qualitative examples in Fig.~\ref{fig:arctic_combined_ablation} show consistent trends, with the full model better preserving object appearance, state, and hand-object configurations across the generated sequence.

\textbf{Multi-View Exocentric Training.}
We further study robustness to source viewpoints by training with one to six exocentric cameras under the same number of optimization steps. As shown in Fig.~\ref{fig:arctic_combined_ablation}, increasing viewpoint diversity generally improves performance, although the trend is not strictly monotonic. PSNR peaks at 14.01\,dB with five views, while six-view training achieves the best LPIPS (0.3860), mIoU (0.3881), and PA-MPJPE (10.18\,mm). These results suggest that diverse exocentric training viewpoints improve generalization to viewpoint changes, with performance largely saturating around five to six views.

\section{Conclusion}
We present \method{}, an HOI-aware framework for exocentric-to-egocentric video generation that aims to faithfully preserve demonstrated hand-object interactions across large viewpoint changes. Our key idea is to complement scene-level geometric conditioning with explicit interaction guidance and object-centric anchoring. The unified 4D HOI prior jointly models scene structure, articulated hands, and hand-object relations to preserve fine-grained interaction, while Decomposed Gated Cross-Attention anchors object-specific semantic and appearance cues throughout egocentric video generation to preserve object consistency across viewpoints. 
Experiments on ARCTIC-HOI and Ego-Exo4D show that Exo2EgoHOI better preserves manipulated objects and hand configurations under large viewpoint changes. On ARCTIC-HOI, it improves object mIoU by 32.3\% and reduces PA-MPJPE by 50.0\% relative to the respective best baselines. 

\noindent\textbf{Limitations and future work.}
Our framework relies on estimated depth, hand geometry, and object segmentation, whose errors may propagate into the reconstructed 4D HOI prior and affect generation quality. Future work will explore more robust interaction reconstruction and the use of generated egocentric videos as human demonstrations for downstream embodied learning and robot manipulation.

\subsection*{AI Use Statement}
ChatGPT and Codex were used during manuscript preparation to support language editing, including grammar correction, sentence refinement, readability improvement, and consistency of technical terminology. All AI-assisted edits were carefully reviewed and revised by the authors to ensure that they faithfully reflect the intended scientific claims, methodology, and experimental results. The authors assume full responsibility for the final manuscript.

\subsection*{Ethics Statement}
The authors acknowledge and adhere to the ICLR Code of Ethics. The research presented in this paper, as well as our participation in the submission and review process, follows the ethical principles and professional standards established by ICLR.

\subsection*{Reproducibility statement}
To ensure reproducibility and verification of our work, we include the implementation details and evaluation procedures in our appendix and will make our source code publicly available upon acceptance.

\bibliography{ref}
\bibliographystyle{iclr2027_conference}

\appendix
\section{Appendix}

This supplementary material provides additional implementation, evaluation, and qualitative details for \method{}. We first describe the model architecture, HOI adapter, global-local anchor construction, and training and inference settings. We then detail the construction of ARCTIC-HOI and the preprocessing pipeline for scene geometry, object masks, and hand reconstruction. Next, we provide complete definitions of the visual fidelity, object consistency, and hand consistency metrics used in the main paper. Finally, we present additional qualitative comparisons on ARCTIC-HOI and Ego-Exo4D, followed by representative failure cases and an analysis of the current limitations.

\subsection{Architecture and Implementation Details}
\label{sec:supp_implementation}

\textbf{Backbone and latent layout.}
We use Wan2.1-I2V-14B-480P~\citep{wan} as our backbone, which contains 40 transformer blocks, hidden width 5120, 40 attention heads, and latent patch size $(1,2,2)$. Our 36 input channels comprise the current video latent (dim=16), conditioning mask (dim=4), and visual condition (dim=16).
The frozen causal Wan VAE has encoder widths $(96,192,384,384)$ and two
residual blocks per stage. Spatial compression is eight; temporal
compression maps $T$ frames to $(T-1)/4+1$. Exocentric, ego-target, and rendered scene-prior videos are encoded separately before widthwise latent concatenation. Exocentric $448\times784$ and ego $448\times448$ images produce a combined $13\times56\times154$ latent grid, or $13\times28\times77$ transformer tokens. Only ego latents receive noise and supervision during the training process.

\begin{table}[h]
\centering
\small
\setlength{\tabcolsep}{4pt}
\begin{tabular}{lcccc}
\toprule
Operation & Channels & Kernel & Stride & Output $(T,H,W)$ \\
\midrule
Branch Conv 1--3 & $C\to16\to16\to16$ & $(3,3,3)$ & $(1,1,1)$ & $(49,448,448)$ \\
Branch Conv 4 & $16\to16$ & $(3,3,3)$ & $(1,2,2)$ & $(49,224,224)$ \\
Branch Conv 5 & $16\to16$ & $(3,3,3)$ & $(2,2,2)$ & $(25,112,112)$ \\
Branch Conv 6 & $16\to16$ & $(3,3,3)$ & $(2,2,2)$ & $(13,56,56)$ \\
Concatenate + fuse & $16+16\to16$ & $(1,1,1)$ & $(1,1,1)$ & $(13,56,56)$ \\
Patch projection & $16\to5120$ & $(1,2,2)$ & $(1,2,2)$ & $(13,28,28)$ \\
\bottomrule
\end{tabular}
\caption{HOI adapter layers; $C=3$ for hand mesh prior and $C=5$ for interaction prior. Branch convolutions use padding one and SiLU, fusion and projection use zero padding.}
\label{tab:supp_adapter}
\end{table}

\textbf{Dual-branch HOI adapter.}
Independent branches encode hand prior and the five-channel interaction prior directly in pixel space (see Tab.~\ref{tab:supp_adapter} for detailed architecture). Hand prior retains $[-1,1]$ scaling and the interaction prior retains its geometric normalization.
Every branch convolution has bias and SiLU~\citep{silu}, with symmetric padding. The final projection's weights and biases are zero-initialized. The resulting residual is initially zero and is added once to the ego token grid immediately after input patchification. 

\textbf{Global-Local Anchor Construction.}
Frozen CLIP-ViT-H/14~\citep{clip} resizes images to $224^2$ without center cropping and supplies penultimate-layer features: one $[\texttt{CLS}]$ plus 256 patch tokens, each 1280-dimensional. 
The object crop expands the tight mask box by 20\% on each side and pads to a square with RGB value 0.5. The background image replaces object pixels with 0.5. 
Their $[\texttt{CLS}]$ tokens form the global reference. 
A shared pretrained projector maps features to width 5120.
Each block shares image key/value projections across global and local contexts.

\textbf{Training and inference settings.}
We jointly train the LoRA parameters, adapter, and gates, while freezing all other pretrained weights. 
LoRA adapts the query, key, value, and output projections in self- and cross-attention, with rank and alpha both set to 256. 
The flow loss uses timesteps sampled uniformly from a 1000-step Euler training schedule and is computed as the egocentric latent velocity MSE. 
We use AdamW with learning rate $2\times10^{-5}$, betas $(0.9,0.95)$, $\epsilon=10^{-8}$, and weight decay $10^{-4}$. The learning rate remains constant after 100 warm-up steps, and the gradient clipping threshold is 1. Training uses four processes, a per-process batch size of 1, gradient accumulation of 1, and seed 42. 
Inference uses second-order UniPC~\citep{zhao2023unipc} with flow shift 3, 50 sampling steps, and guidance scale 5 to generate 49-frame videos at 30 fps.

\subsection{ARCTIC-HOI Benchmark Construction}
\label{sec:supp_data}
\begin{figure*}[t]
  \centering
  \includegraphics[width=0.89\textwidth]{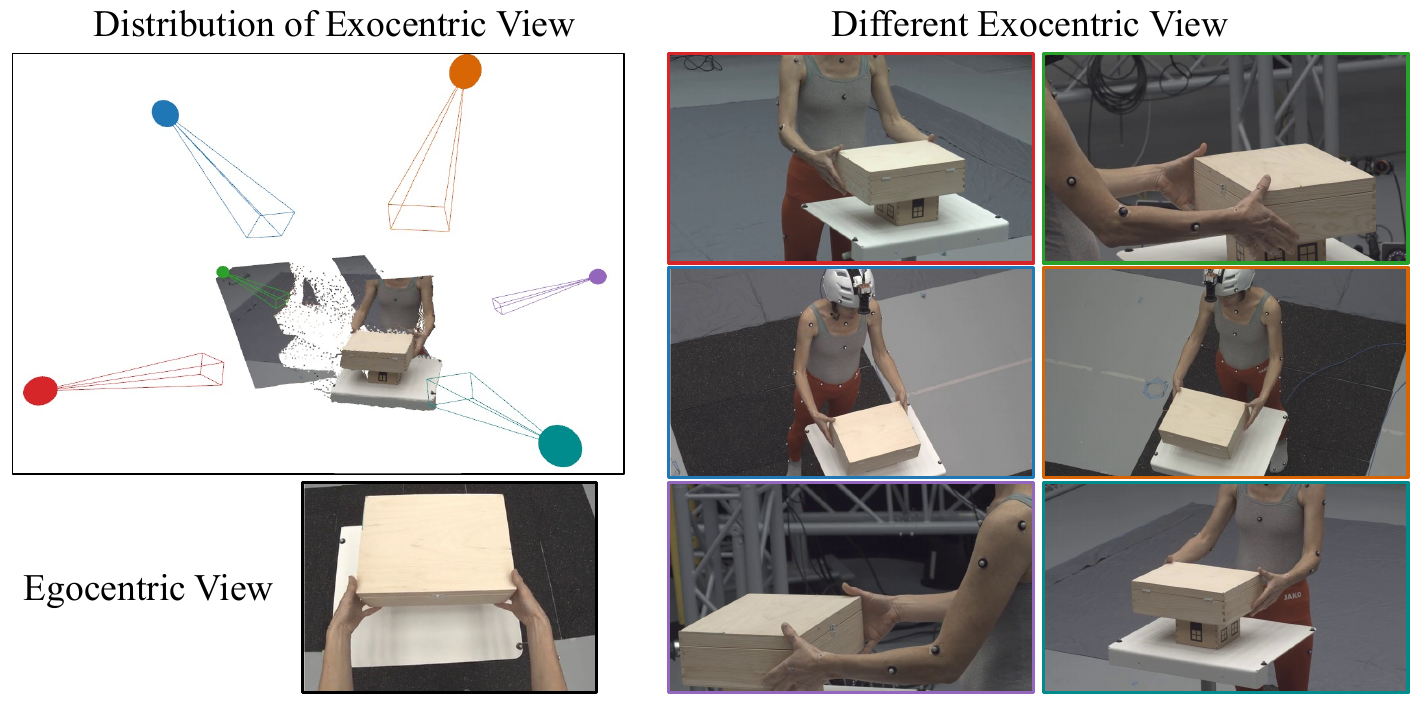}
  \caption{
  Left: reconstructed 3D scene and camera configuration, with the corresponding egocentric observation shown below. Colored camera frustums denote different exocentric viewpoints. Right: corresponding images of the same scene observed from these exocentric cameras.}
\label{fig:supp_dataset}
\end{figure*}

\textbf{ARCTIC-HOI Dataset Construction.}
As illustrated in Fig.~\ref{fig:supp_dataset}, we construct paired exocentric--egocentric video clips from the ARCTIC~\citep{arctic} sequences. We designate camera 0 as the egocentric view and cameras $\{1,3,4,5,6,8\}$ as exocentric views. Cameras 2 and 7 are excluded because their viewpoints deviate substantially from the designated egocentric view. After removing the first and last 40 frames of each sequence, we independently sample four non-overlapping 49-frame intervals for each exocentric-egocentric camera pair. This procedure yields approximately 7,000 training clips and 180 test clips, with 30 test clips for each exocentric viewpoint.
To further evaluate robustness in unconstrained scenarios, we additionally collect 60 paired exocentric--egocentric clips from in-the-wild human manipulation videos. For these sequences, the relative camera poses between the first- and third-person views are calibrated using AprilTag~\citep{olson2011tags} placed on the tabletop. These videos cover more diverse environments, objects, and manipulation patterns beyond the controlled ARCTIC setting. Combining the ARCTIC-derived split with these in-the-wild sequences, our ARCTIC-HOI benchmark contains approximately 7,000 training clips and 240 test clips in total.
We use Depth Anything~3 (DA3)~\citep{depthanything3}
(\texttt{DA3NESTED-GIANT-LARGE-1.1}) to reconstruct scene geometry offline.
Importantly, no egocentric RGB images are used for depth estimation. Our video generation takes only a single exocentric clip together with the target camera trajectory. For each selected source view, we back-project its RGB-D observation into a point cloud and render only this source-view point cloud into the target
egocentric camera. Rendering uses a depth buffer, point size 7, and
$2\times$ supersampling, with uncovered regions set to black.

\textbf{Object Masks and Hand Reconstruction.}
We obtain object and hand masks using GroundingDINO Swin-T OGC
~\citep{liu2024grounding_dino} followed by SAM~2.1 Hiera-Large~\citep{sam2}.
Objects and hands are detected separately using the prompts \texttt{object.} and \texttt{hand.}, with box and text thresholds of 0.35 and 0.25, respectively. Masks detected under the same prompt are merged, and segmentation is performed independently for each frame without temporal mask propagation.
For hand reconstruction, WiLoR~\citep{wilor} predicts 778-vertex
MANO~\citep{mano} meshes with a confidence threshold of 0.3, an IoU
threshold of 0.5, and crop rescaling factor 2.0. We align each reconstructed
hand mesh to the observed scene depth using valid depth measurements within
the corresponding hand mask. Depth alignment preserves the estimated hand articulation, orientation, and scale, while hands without sufficient depth support are discarded. Finally, egocentric hand priors are rendered using fixed colors for the left and right hands
with hand-only visibility, without applying object-depth occlusion.

\subsection{Evaluation metrics}
\label{sec:supp-evaluation}

\subsubsection{Visual fidelity}
We evaluate visual fidelity using PSNR, SSIM, LPIPS, and CLIP-I, with scores computed per frame before averaging. PSNR and SSIM use RGB images with a data range of 255 and uniform $7\times7$ windows for SSIM. LPIPS uses pretrained AlexNet, with inputs normalized to $[-1,1]$ and no additional resizing. CLIP-I measures cosine similarity between $\ell_2$-normalized image features from \path{openai/clip-vit-base-patch32}. Its preprocessing consists of bicubic short-side resizing to 224, and a $224\times224$ center crop.

\subsubsection{Object consistency}
\label{sec:supp-object-evaluation}
We independently extract object masks from each reference and generated frame using Grounding DINO Swin-T OGC~\citep{liu2024grounding_dino} and the SAM~2.1 Hiera-L image predictor~\citep{sam2}. We use the prompt \texttt{object.}, box and text thresholds of 0.35 and 0.25, and single-mask SAM output. Detector inputs are resized to a short side of 800 with a maximum side of 1333 and use ImageNet normalization. 

\textbf{mIoU and Center-Err.}
For reference and predicted masks $M_t,\widehat M_t$, let $\mathcal G$ contain frames with a reference mask and $\mathcal B$ contain frames where both masks are present. We compute the following per-video scores:
\begin{equation}
 \operatorname{mIoU}=\frac{1}{|\mathcal G|}\sum_{t\in\mathcal G}
 \frac{|\widehat M_t\cap M_t|}{|\widehat M_t\cup M_t|},\qquad
 \operatorname{Center\mbox{-}Err.}=\frac{1}{|\mathcal B|}
 \sum_{t\in\mathcal B}\frac{\|c(\widehat M_t)-c(M_t)\|_2}
 {\sqrt{H^2+W^2}},
\end{equation}
where $c(M)$ is the foreground-pixel centroid. mIoU measures mask overlap over reference-present frames, assigning zero to missing predictions and excluding reference-absent frames. Center-Err.\ measures centroid displacement normalized by the full image diagonal, including padding, and is evaluated only when both masks are present.

\emph{CLIP-O.}
CLIP-O measures object appearance consistency by averaging CLIP image-feature cosine similarity over $\mathcal B$, using the same encoder and preprocessing as CLIP-I. Object crops are constructed independently from the reference and predicted masks. Each mask's tight bounding box is expanded by $\lceil0.2w\rceil$ on each horizontal side and $\lceil0.2h\rceil$ on each vertical side, where $w$ and $h$ are the box width and height. The expanded crop is clipped to the image and centered on a square canvas. Non-mask pixels and square padding are filled with RGB $(127,127,127)$. Both CLIP-O and Center-Err.\ are conditional on both masks being present and do not penalize missing predictions.

\subsubsection{Hand consistency}
\label{sec:supp-hand-evaluation}
We use pretrained WiLoR~\citep{wilor} to independently estimate 21 hand joints from reference and generated RGB frames. 
We use the WiLoR detector with confidence threshold 0.3, NMS IoU threshold 0.5, box expansion factor 2.0, and batch size 32. Pre-processing uses normalized $256^2$ hand crops and left-hand flipping followed by restoration.
For each frame and reference side, we select the highest-confidence reference detection and the highest-confidence same-side prediction. If no same-side prediction exists, we select the hand with the nearest camera-space wrist among all predicted hands. Reference-missing observations are excluded.

\emph{Error definition.}
For associated joints $\widehat{\mathbf p}_{t,s,j},\mathbf p_{t,s,j}$ in millimeter, we define
\begin{equation}
 e_{t,s}(A)=\frac{1}{21}\sum_{j=1}^{21}
 \|A(\widehat{\mathbf p}_{t,s,j})-\mathbf p_{t,s,j}\|_2,
 \qquad A(\mathbf p)=aR\mathbf p+\mathbf b.
 \label{eq:supp_hand_error}
\end{equation}
The three hand metrics differ in their alignment $A$:

\begin{itemize}
  \item \emph{MPJPE} uses the identity transformation $A$ and measures joint errors directly in camera coordinates, without wrist centering.
  \item \emph{PA-MPJPE} fits a separate similarity transformation $A$ for each frame and reference side, allowing scale, rotation, and translation alignment independently at each frame.
  \item \emph{WA-MPJPE} fits one similarity transformation $A$ per video and reference side using all associated frames, including wrist-fallback matches. The same transformation is applied throughout the sequence. This aligns WiLoR camera-coordinate sequences and does not recover a physical world coordinate system.
\end{itemize}

For PA-MPJPE and WA-MPJPE, alignment minimizes squared joint residuals via SVD over scale, proper rotation without reflection, and translation. The reported errors in Eq.~\ref{eq:supp_hand_error} are mean Euclidean distances in millimeters.

\subsection{Additional Qualitative Results}
\label{sec:supp-additional-results}

We provide additional qualitative results to further evaluate \method{} across different datasets, viewpoints, and manipulation scenarios. In addition to visual comparisons of the generated egocentric videos, we analyze how well different methods preserve the demonstrated hand configurations and manipulated objects under large exocentric-to-egocentric viewpoint changes.

\begin{figure*}[t]
  \centering
  \includegraphics[width=0.99\textwidth]{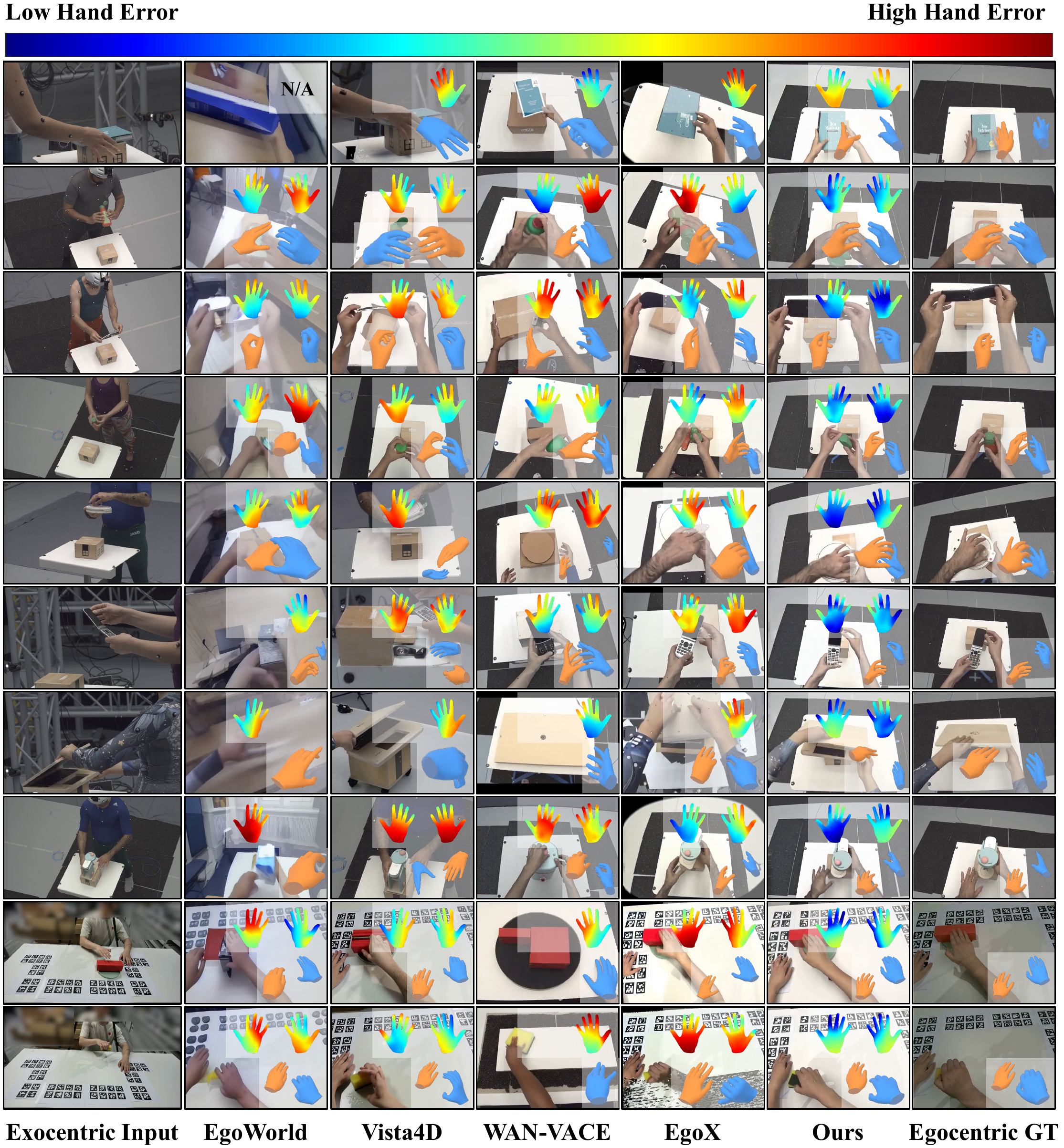}
\caption{\textbf{Additional qualitative comparison on ARCTIC-HOI.}
We compare egocentric videos generated by different methods and overlay their reconstructed 3D hand meshes. Hand-mesh vertices are color-coded by errors relative to the egocentric ground truth, ranging from blue (low error) to red (high error); \textit{N/A} indicates failed hand reconstruction. The visualization highlights differences in target-view hand configuration and hand-object alignment across methods.}
\label{fig:supp_arctic}
\end{figure*}

On ARCTIC-HOI (see Fig.~\ref{fig:supp_arctic}), we provide additional comparisons with existing methods and overlay the reconstructed 3D hand meshes on the generated egocentric frames. We further color-code the per-vertex hand errors with respect to the egocentric ground truth, providing a more direct visualization of hand-pose consistency beyond image-level appearance. These examples cover diverse hand configurations, object geometries, and bimanual interactions, and complement the quantitative hand-consistency metrics reported in the main paper.

\begin{figure*}[t]
  \centering
  \includegraphics[width=0.99\textwidth]{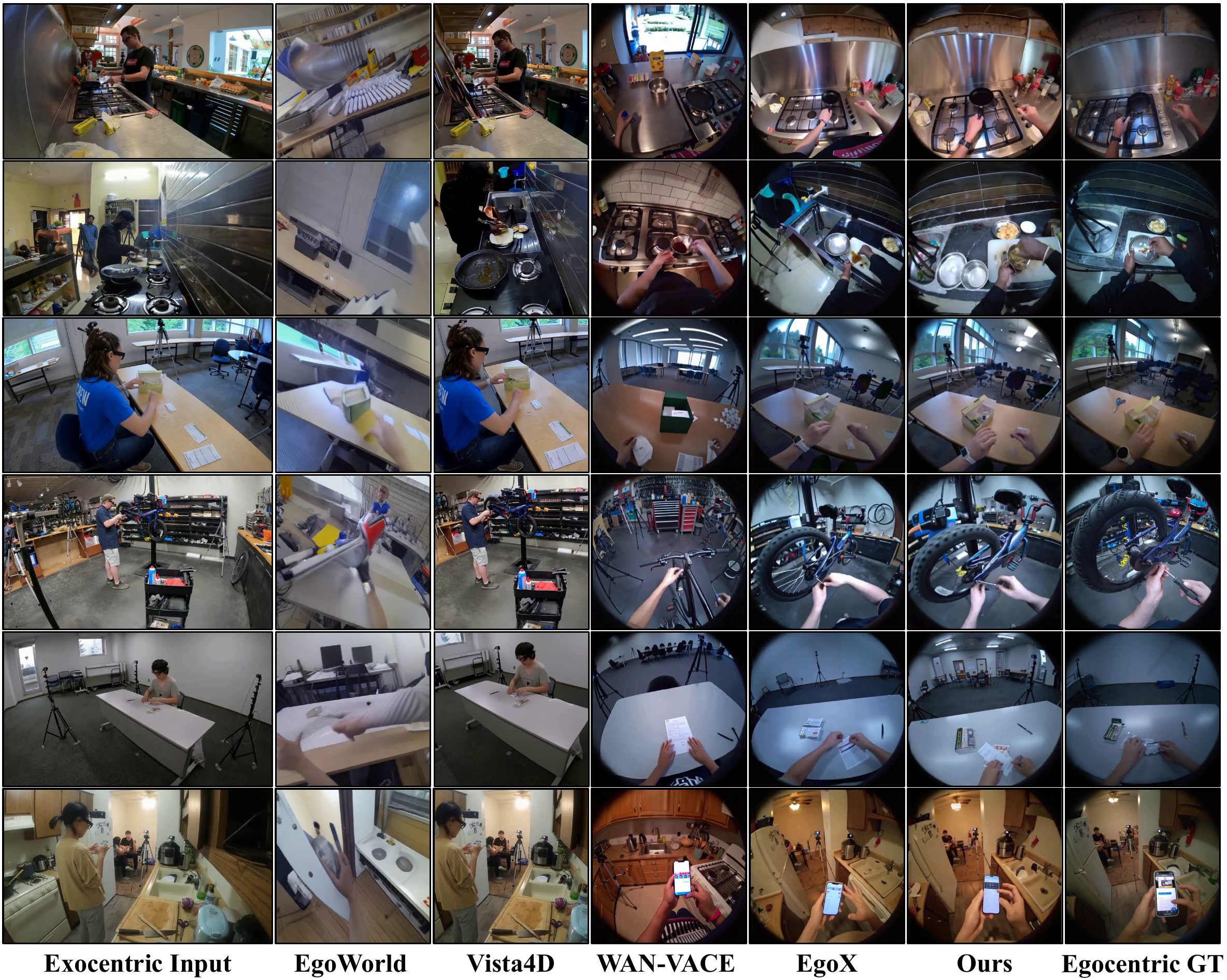}
  \vspace{-3mm}
\caption{\textbf{Additional qualitative comparison on Ego-Exo4D.}
We compare egocentric videos generated from exocentric inputs across diverse real-world scenes and manipulation activities. The results illustrate differences among methods in cross-view scene reconstruction, manipulated-object consistency, and hand-object configurations under substantial viewpoint changes.}
\vspace{-3mm}
\label{fig:supp_egoexo4d}
\end{figure*}

We also present additional results on Ego-Exo4D (see Fig.~\ref{fig:supp_egoexo4d}), which contains more diverse real-world environments, camera configurations, and manipulation activities. These examples provide a complementary evaluation of cross-view generation beyond the relatively controlled ARCTIC-HOI setting, particularly in terms of scene reconstruction, manipulated-object consistency, and hand-object configurations.

\begin{figure*}[h]
  \centering
  \includegraphics[width=0.86\textwidth]{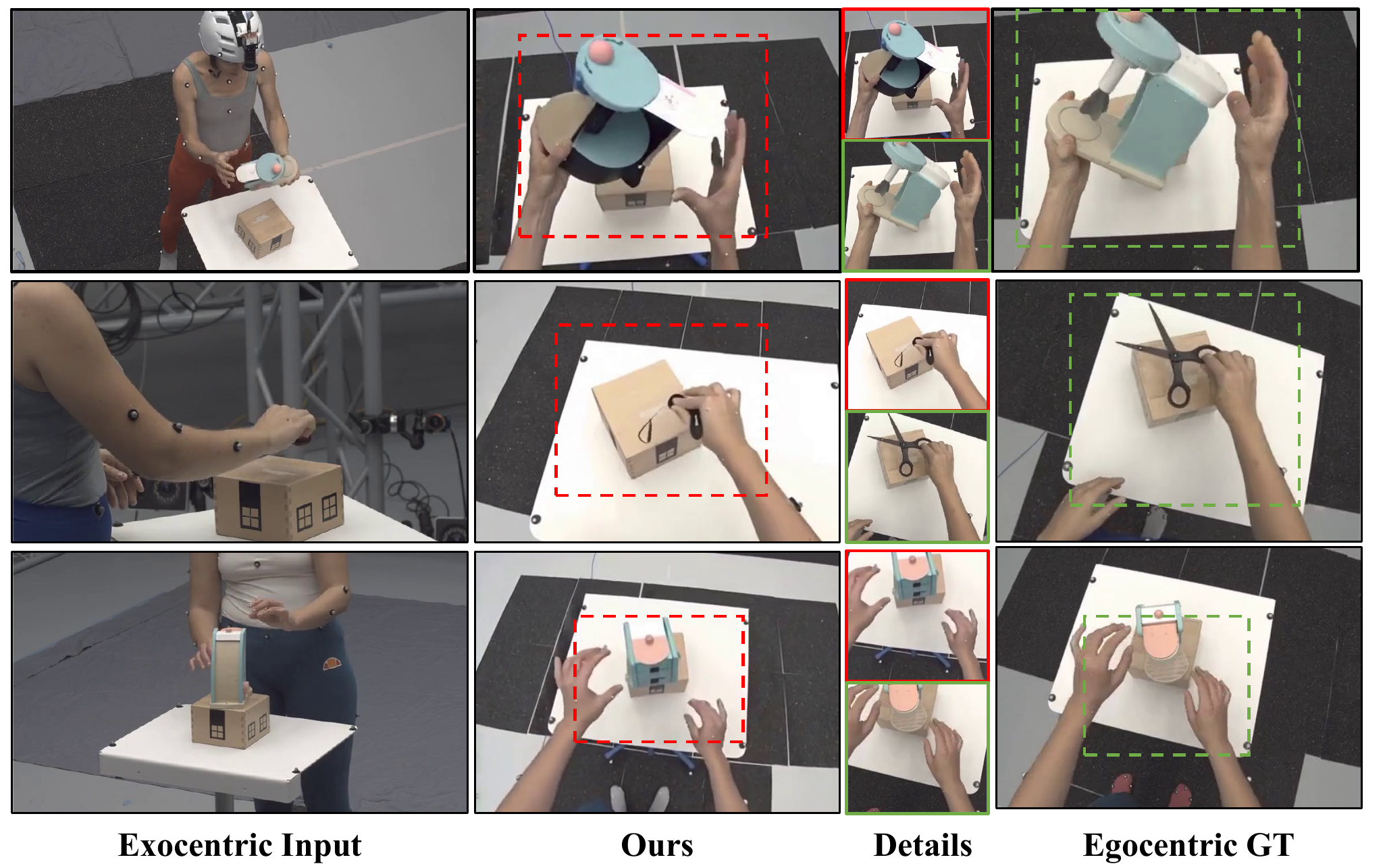}
  \vspace{-3mm}
\caption{\textbf{Representative failure cases of \method{}.}
Severe occlusion can lead to incomplete hand-object reconstruction, while complex articulated or thin-structured objects remain difficult to model accurately. Insets highlight discrepancies in object geometry, articulation state, and hand-object configuration compared with the egocentric ground truth.}
\label{fig:failure_case}
\end{figure*}

Finally, we show representative failure cases (see Fig.~\ref {fig:failure_case}) of \method{}. Severe occlusion of the hands or manipulated objects can provide insufficient geometric evidence for reconstructing the underlying interaction, resulting in inaccurate target-view generation. Complex articulated or thin-structured objects, such as scissors, are also challenging because their fine-scale geometry and articulation states are difficult to recover reliably from exocentric observations. These cases highlight the dependence of our framework on accurate upstream geometry and interaction reconstruction, and motivate more robust HOI reconstruction for future work.

\end{document}